\documentclass[10pt,twocolumn,letterpaper]{article}

\usepackage{cvpr}              
\usepackage{pifont}
\usepackage{makecell}
\usepackage{booktabs}
\usepackage{multirow}
\usepackage{amsmath}
\usepackage{amssymb}
\usepackage{graphicx}
\usepackage{xcolor}
\usepackage{soul} 
\usepackage{algorithm}
\usepackage[noend]{algorithmic}

\definecolor{cvprblue}{rgb}{0.21,0.49,0.74}
\usepackage[pagebackref,breaklinks,colorlinks,allcolors=cvprblue]{hyperref}

\def\paperID{13237} 
\def\confName{CVPR}
\def\confYear{2025}

\title{FedBiP: Heterogeneous One-Shot Federated Learning with Personalized Latent Diffusion Models}

\author {Haokun Chen\textsuperscript{\rm 1,}\textsuperscript{\rm 2,}\textsuperscript{\rm 5}\thanks{Corresponding to \textit{haokun.chen@siemens.com} and \textit{jindong.gu@outlook.com}}, \   
Hang Li\textsuperscript{\rm 1,}\textsuperscript{\rm 2}, \ 
Yao Zhang\textsuperscript{\rm 1,}\textsuperscript{\rm 5}, \ 
Jinhe Bi\textsuperscript{\rm 1}, \ 
Gengyuan Zhang\textsuperscript{\rm 1,}\textsuperscript{\rm 5}, \  \\
Yueqi Zhang\textsuperscript{\rm 4}, \ 
Philip Torr\textsuperscript{\rm 3}, \ 
Jindong Gu\textsuperscript{\rm 3*}, \ 
Denis Krompass\textsuperscript{\rm 2}, \ 
Volker Tresp\textsuperscript{\rm 1,}\textsuperscript{\rm 5} \\
\textsuperscript{\rm 1} Ludwig Maximilian University of Munich \quad \textsuperscript{\rm 2} Siemens Technology \quad
\textsuperscript{\rm 3} University of Oxford \\  \textsuperscript{\rm 4} Technical University of Munich  \quad \textsuperscript{\rm 5} Munich Center for Machine Learning 
}

\begin{document}
\maketitle
\begin{abstract}
One-Shot Federated Learning (OSFL), a special decentralized machine learning paradigm, has recently gained significant attention. OSFL requires only a single round of client data or model upload, which reduces communication costs and mitigates privacy threats compared to traditional FL. Despite these promising prospects, existing methods face challenges due to client data heterogeneity and limited data quantity when applied to real-world OSFL systems. Recently, Latent Diffusion Models (LDM) have shown remarkable advancements in synthesizing high-quality images through pretraining on large-scale datasets, thereby presenting a potential solution to overcome these issues. However, directly applying pretrained LDM to heterogeneous OSFL results in significant distribution shifts in synthetic data, leading to performance degradation in classification models trained on such data. This issue is particularly pronounced in rare domains, such as medical imaging, which are underrepresented in LDM's pretraining data. To address this challenge, we propose Federated Bi-Level Personalization (FedBiP), which personalizes the pretrained LDM at both instance-level and concept-level. Hereby, FedBiP synthesizes images following the client's local data distribution without compromising the privacy regulations. FedBiP is also the first approach to simultaneously address feature space heterogeneity and client data scarcity in OSFL. Our method is validated through extensive experiments on three OSFL benchmarks with feature space heterogeneity, as well as on challenging medical and satellite image datasets with label heterogeneity. The results demonstrate the effectiveness of FedBiP, which substantially outperforms other OSFL methods. Our code is available at \href{https://github.com/HaokunChen245/FedBiP}{https://github.com/HaokunChen245/FedBiP}.
\end{abstract}    
\section{Introduction}

Federated Learning (FL) \citep{mcmahan2017communication} is a decentralized machine learning paradigm, in which multiple clients collaboratively train neural networks without centralizing their local data. However, traditional FL frameworks require frequent communication between a server and clients to transmit model weights, which would lead to significant communication overheads \citep{kairouz2021advances}. Additionally, such frequent communication increases system susceptibility to privacy threats, as transmitted data can be intercepted by attackers who may then execute membership inference attacks \citep{lyu2020threats}. In contrast, a special variant of FL, One-Shot Federated Learning (OSFL) \citep{guha2019one}, serves as a promising solution. OSFL requires only single-round server-client communication, thereby enhancing communication efficiency and significantly reducing the risk of interception by malicious attackers. Therefore, we focus on OSFL given its promising properties. 

\begin{figure}
    \centering
    \includegraphics[width=\linewidth]{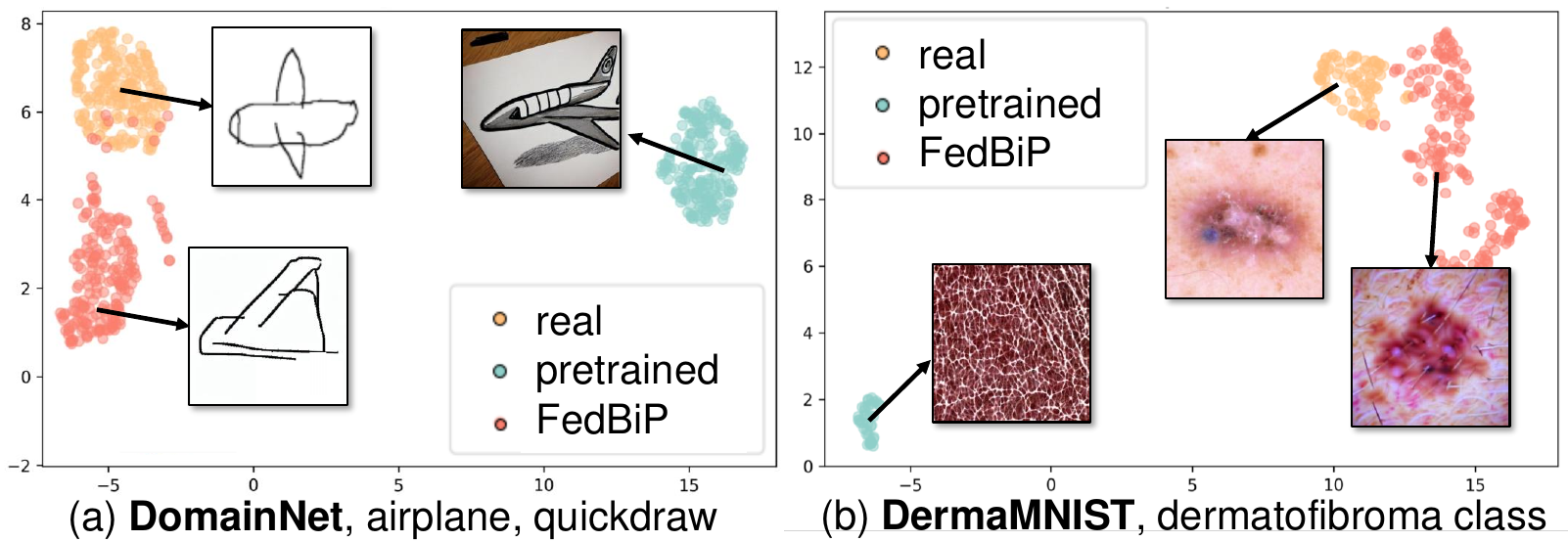}
    \vspace{-12pt}
    \caption{Feature map visualization of client images (\textit{real}), synthetic images by prompted pretrained LDM (\textit{pretrained}), and our method (\textit{FedBiP}) on two datasets. \texttt{FedBiP} mitigates the strong distribution shifts between pretrained LDM and client local data.}
    \vspace{-15pt}
    \label{fig:motivation}
\end{figure}

Despite these promising prospects, existing methods for OSFL face significant challenges when applied to real-world scenarios. Previous works \citep{guha2019one, li2020practical} require additional public datasets, presenting challenges in privacy-critical domains such as medical data \citep{liu2021feddg}, where acquiring data that conforms to client-specific distributions is often impractical. Alternatively, they can involve the transmission of entire model weights \citep{zhang2022dense} or local training data \citep{zhou2020distilled}, which are inefficient and increase the risk of privacy leakage. Moreover, these approaches overlook the issue of feature space heterogeneity, wherein the data features across different clients exhibit non-identically distributed properties. This presents an important and prevalent challenge as emphasized in \citep{li2021fedbn, chen2023fraug}. Another vital challenge in (One-Shot) FL is the limited quantity of data available from clients \citep{mcmahan2017communication}. This problem is particularly notable in specialized domains, such as medical or satellite imaging \citep{so2022fedspace} where data collection is time-consuming and costly.

Data augmentation constitutes a promising strategy to address these challenges in traditional FL \citep{zhu2021data, li2022federated} by optimizing an auxiliary generative model. However, its reliance on multiple communication rounds makes it unsuitable for OSFL. Recently, diffusion models \citep{ho2020denoising}, particularly Latent Diffusion Model (LDM) \citep{rombach2022high}, have gained significant attention due to their capability to synthesize high-quality images after being pretrained on large-scale datasets. They are proven effective in various tasks, including training data augmentation \citep{yuan2023real, azizi2023synthetic} and addressing feature shift problems \citep{niemeijer2024generalization, gong2023prompting} under centralized settings. However, directly applying a pretrained LDM for specialized domains presents challenges. As demonstrated in Figure \ref{fig:motivation}, there is a noticeable feature distributional shift and visual discrepancy between real and synthetic data. This mismatch could lead to performance degradation when incorporating such synthetic data into training, especially in heterogeneous OSFL, where each client possesses data with varying distributions.


Therefore, in this paper, we propose Federated Bi-Level Personalization (\texttt{FedBiP}), a framework designed to adapt pretrained LDM for synthesizing high-quality training data that adheres to client-specific data distributions in OSFL. \texttt{FedBiP} incorporates personalization of the pretrained LDM at both instance and concept levels. Specifically, instance-level personalization focuses on adapting the pretrained LDM to generate high-fidelity samples that closely align with each client's local data while preserving data privacy. Concurrently, concept-level personalization integrates category and domain-specific concepts from different clients to enhance data generation diversity at the central server. This bi-level personalization approach improves the classification models trained on the synthesized data. Our contributions can be summarized as follows:
\begin{itemize}
    \item We propose a novel method \texttt{FedBiP} to incorporate pretrained Latent Diffusion Model (LDM) for heterogeneous OSFL, marking the first OSFL framework to tackle feature space heterogeneity via personalizing LDM.
    \item We conduct comprehensive experiments on three OSFL benchmarks with feature space heterogeneity, in which \texttt{FedBiP} achieves state-of-the-art results.
    \item We validate the maturity and scalability of \texttt{FedBiP} on real-world medical and satellite image datasets with label space heterogeneity, and further demonstrate its promising capability in preserving client privacy.
\end{itemize} 

\section{Related Works}
\subsection{One-Shot Federated Learning}
A variety of efforts have been made to address One-Shot Federated Learning (OSFL), primarily from two complementary perspectives: one focuses on model aggregation through techniques such as model prediction averaging \citep{guha2019one}, majority voting \citep{li2020practical}, conformal prediction method \citep{humbert2023one}, loss surface adaptation \citep{su2023one}, or Bayesian methods \citep{yurochkin2019bayesian, chen2020fedbe, hasan2024calibrated}. These approaches may not fully exploit the underlying knowledge across different client data distributions. Another aims to transmit training data instead of model weights: data distribution \citep{kasturi2020fusion, beitollahi2024parametric, shin2020xor}, Generative Adversarial Networks (GANs) \citep{goodfellow2020generative, zhang2022dense, kasturi2023osgan, kang2023one, dai2024enhancing}, or distilled dataset \citep{zhou2020distilled, song2023federated} are optimized and transmitted to the central server for subsequent model training. Given the success of diffusion models \citep{rombach2022high}, \citep{zhang2023federated, yang2024feddeo} suggests transmitting image captions to reproduce training data at the server, while \citep{yang2024exploring} focuses on one-shot semi-supervised FL. However, these approaches are either inefficient or pose risks of client information leakage. In contrast, \texttt{FedBiP} functions as an OSFL algorithm, offering enhanced efficiency and robust privacy-preserving capabilities.


\subsection{Diffusion Models For Image Synthesis}
Diffusion models \citep{ho2020denoising}, especially Latent Diffusion Model (LDM) \citep{rombach2022high}, have attracted significant attention due to their capability to generate high-resolution natural images. They have demonstrated effectiveness in various applications, including image stylization \citep{guo2023animatediff, meng2021sdedit, kawar2023imagic} and training data generation \citep{yuan2023real, sariyildiz2023fake, azizi2023synthetic}. We refer readers to \citep{croitoru2023diffusion, yang2023diffusion} for a comprehensive overview of recent progress on diffusion models. Pretrained LDM has been adopted to address client data scarcity in OSFL \citep{zhang2023federated, yang2024feddeo}. However, these methods often overlook the feature distribution shift between the LDM pretraining dataset and the clients' local data. This challenge is particularly pronounced in complex domains such as medical and satellite imaging. To address this issue, we propose \texttt{FedBiP}, which personalizes the pretrained LDM to synthesize data aligned with the client data distributions.

\section{Preliminaries}
\subsection{Heterogeneous One-shot Federated Learning}
In this section, we introduce our problem setting, i.e., heterogeneous One-Shot Federated Learning (OSFL). Following \citep{zhang2023federated}, we focus on image classification tasks with the goal of optimizing a $C$-way classification model $\phi$ utilizing the client local data, where $C \in \mathbb{N}$ denotes the number of categories. We assume there are $K \in \mathbb{N}$ clients joining the collaborative training. Each client $k$ owns its private dataset $D^k$ containing $N^k \in \mathbb{N}$ (image, label) pairs: $\{x^k_i, y^k_i\}_{i=1}^{N^k}$. Only one-shot data upload from the clients to the central server is allowed.

As described in \citep{kairouz2021advances}, OSFL with data heterogeneity is characterized by distribution shifts in local datasets: $P_\mathcal{XY}^{k_1} \neq P_\mathcal{XY}^{k_2}$ with $k_1 \neq k_2$, where $P_\mathcal{XY}^{k}$ defines the joint distribution of input space $\mathcal{X}$ and label space $\mathcal{Y}$ on $D^k$. Data heterogeneity can be decomposed into two types: (1) \textit{label space} heterogeneity, where $P_\mathcal{Y}$ varies across clients, while $P_\mathcal{X|Y}$ remains the same, and (2) \textit{feature space} heterogeneity, where $P_\mathcal{X}$ or $P_\mathcal{X|Y}$ varies across clients, while $P_\mathcal{Y|X}$ or $P_\mathcal{Y}$ remains the same.

\subsection{Latent Diffusion Model Pipeline}
In this section, we introduce the training and inference pipelines for Latent Diffusion Model (LDM). Given an image $x \in \mathbb{R}^{H \times W \times 3}$, the encoder $\mathcal{E}$ encodes $x$ into a latent representation $z(0)=\mathcal{E}(x)$, where $z(0) \in \mathbb{R}^{h \times w \times c}$. Besides, the decoder $\mathcal{D}$ reconstructs the image from the latent, giving $\tilde{x} = \mathcal{D}(z(0)) = \mathcal{D}(\mathcal{E}(x))$. The forward diffusion and denoising processes occur in the latent representation space, as described below.


In the forward diffusion of LDM training, random noise $\epsilon \sim \mathcal{N}(0,I)$ is added to $z(0)$, producing 
\begin{equation}
    z(t) = \delta(t, z(0)) = \sqrt{\alpha_t} z(0) + \sqrt{1-\alpha_t} \epsilon,
    \label{eq:delta}
\end{equation}

where $t \sim \text{Uniform}(\{1,...,T\})$ is the timestep controlling the noise scheduler $\alpha_t$. A larger $t$ corresponds to greater noise intensity. In the denoising process, a UNet $\epsilon_\theta$ is applied to denoise $z(t)$, yielding $\tilde{z}(0)$ for image reconstruction. To further condition LDM generation on textual inputs $P$, a feature extractor $\tau_\theta$ is used to encode the prompts into intermediate representations for $\epsilon_\theta$. By sampling different values of $\epsilon$ and $t$, $\epsilon_\theta$ can be optimized via the following loss function:
\begin{equation}
    L_{LDM} = \mathbb{E}_{z(0),P,\epsilon,t}\left[||\epsilon - \epsilon_{\theta}(\delta(t, z(0)), t, \tau_{\theta}(P))||_2^2\right]
\end{equation}
In the inference stage, latent representation $z(T)$ will be sampled directly from $\mathcal{N}(0,I)$, and multiple denoising steps are executed to obtain $\tilde{z}(0)$. The image is then decoded via $\tilde{x} = \mathcal{D}(\tilde{z}(0))$.
\begin{algorithm}[ht]
\caption{Training process of \texttt{FedBiP}}
\small
\label{algo:fedmla}
\textbf{ServerUpdate}
\begin{algorithmic}[1]
\STATE Initialize Latent Diffusion Model with pretrained weights $\theta$, classification model $\phi$, synthetic dataset $D_{syn} \gets \varnothing$\\
\FOR [\textbf{in parallel}]{client $k$ = 1 to $K$} 
    \STATE $k^{th}$ client execute $ClientUpdate(k)$ and upload $\{z^{k}_i(T), y^k_i\}_{i=1}^{N_k}, \{V_j^{k}\}_{j=1}^{C}, S^k$ 
    \FOR {$i$ = 1 to $N_k$} 
        \STATE $e \gets \tau_{\theta}$("A $[S^k]$ style of a $[V^{k}_{y_i^k}]$")
        \STATE $\tilde{z}(0) \gets \epsilon_{\theta}(z^k_i(T), t, e)$, $\tilde{x} \gets \mathcal{D}(\tilde{z}(0))$
        \STATE $D_{syn}.append([\tilde{x}, y^k_i])$
    \ENDFOR
\ENDFOR
\STATE Optimize $\phi$ using $D_{syn}$ (Equation \ref{eq:cls})
\end{algorithmic} 

\textbf{ClientUpdate}$(k)$
\begin{algorithmic}[1]
\STATE Initialize Latent Diffusion Model with pretrained weights $\theta$, randomly initialize $\{V^k_j\}_{j=1}^{C}, S^k$.\\
\FOR {$i$ = 1 to $N^k$}
    \STATE Randomly sample an image $x^k_{i'}$ with $i \neq i', y_i=y_{i'}$
    \STATE $\overline{z}(0) \gets \gamma \mathcal{E}(x^k_i) + (1-\gamma) \mathcal{E}(x^k_{i'})$
    \STATE $z^{k}_i(T) \gets \delta(T, \overline{z}(0))$ 
\ENDFOR
\FOR {local step $st$ = 1 to $N_{step}$} 
    \STATE Sample one mini-batch $\{x^k_b, y^k_b\}$ from $D^k$, timestep $t$ \\
    \STATE $e \gets \boldsymbol{\tau}$(\{''A $[S^k]$ style of a $[V^{k}_{y^k_b}]$"\})
    \STATE Optimize $S^k, \{V^k_j\}_{j=1}^{C}$ (Equation \ref{eq:main})
\ENDFOR
\end{algorithmic}
\end{algorithm}

\section{Methodology}
\subsection{Motivational Case Study}

To substantiate the necessity of the proposed method, we present an empirical analysis to address the following research question: \textit{Can pretrained Latent Diffusion Model (LDM) generate images that are infrequently represented in the pretraining dataset using solely textual conditioning?} Specifically, we adopt two datasets, namely DomainNet \citep{peng2019moment} and DermaMNIST \citep{medmnistv2}, which contain images indicating different styles and images from challenging medical domains, respectively. We prompt LDM with \textit{''A quickdraw style of an airplane.''} to generate airplane images in quickdraw style for DomainNet dataset, and \textit{''A dermatoscopic image of a dermatofibroma, a type of pigmented skin lesions.''} for DermaMNIST. We synthesize 100 images for each setting and adopt a pretrained ResNet-18 \citep{he2016deep} to acquire the feature embeddings of real and synthetic images. Finally, we visualize them using UMAP \citep{mcinnes2018umap}. 

As shown in Figure \ref{fig:motivation}, we observe markedly different visual characteristics between synthetic and real images. Specifically, for DomainNet, there exist significant discrepancies between the "quickdraw" concept demonstration in the original dataset and the pretrained LDM. For DermaMNIST, the pretrained LDM is only able to perceive the general concepts of "dermatoscopic" and "skin lesion", failing to capture category-specific information. This further highlights the difficulties in reproducing medical domain data via LDM. Additionally, there is a substantial gap in the extracted feature embeddings between real and synthetic images. Most importantly, despite the high visual quality of the synthetic images, they may not contribute to the final performance of the classification model. As demonstrated by our experimental results (Table \ref{tab:ablation}), directly applying such prompts to generate images for server-side training sometimes yields worse results than baseline methods. Therefore, it is essential to design a more sophisticated method to effectively personalize the pretrained LDM to the specific domains of client local datasets. These observations motivate our proposed method \texttt{FedBiP}, which mitigates the distribution shifts between pretrained LDM and client local data. We introduce \texttt{FedBiP} in the following.

\begin{figure*}[t]
\centering
\vspace{-15pt}
\includegraphics[scale=0.38]{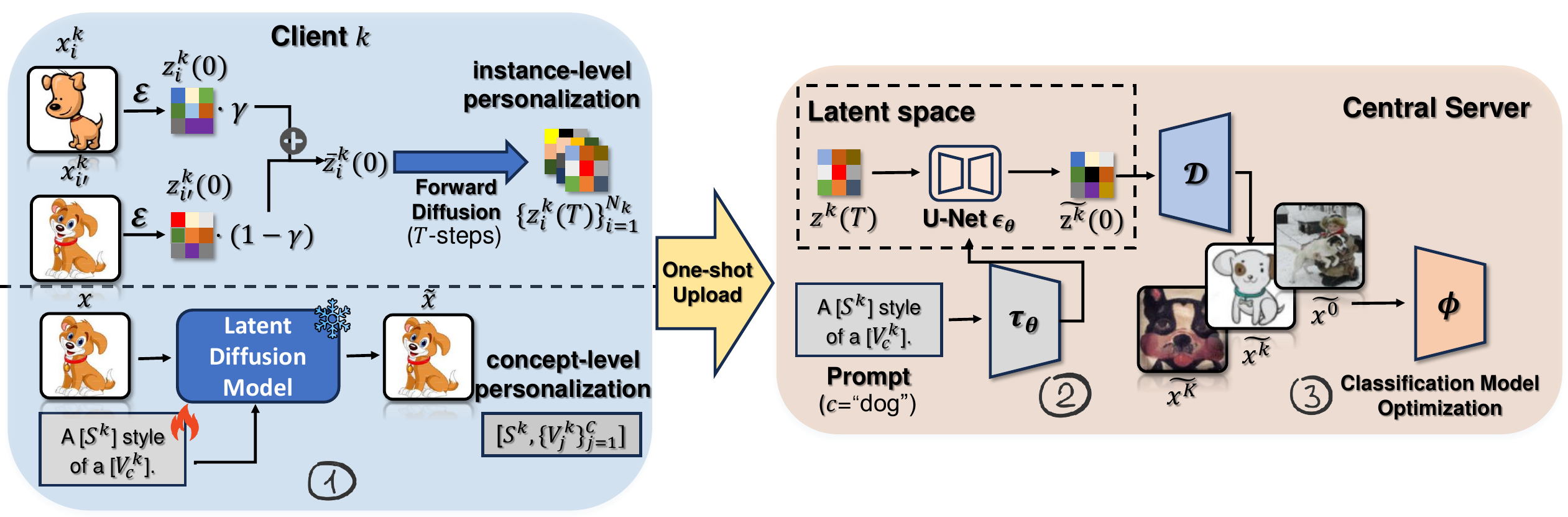} 
\vspace{-8pt}
\caption{Schematic illustration of Federated Bi-Level Personalization (\texttt{FedBiP}). (\ding{172}) Each client executes bi-level personalization and obtains latent vectors $z^k(T)$ and concept vectors $S^k, V^k$. (\ding{173}) The central server integrates the vectors into the generation process of the pretrained Latent Diffusion Model $\theta$. (\ding{174}) The classification model $\phi$ is optimized using synthetic images.}
\vspace{-10pt}
\label{fig:fedmla} 
\end{figure*}

\subsection{Proposed Method}
A schematic overview of the proposed method is provided in Figure \ref{fig:fedmla}. Additionally, the pseudo-code of the proposed method is presented in Algorithm \ref{algo:fedmla}. We begin by introducing the bi-level personalization in the local update of $k^{th}$ client, omitting the subscript $k$ for simplicity in the following description.

\subsubsection{Instance-level Personalization}
\vspace{-2pt}
While the traditional Latent Diffusion Model (LDM) employs a Gaussian distribution to initialize the latent vector $z(T) \sim \mathcal{N}(0, I)$, we directly compute $z(T)$ from the local training set $D^k$ of each client. Specifically, we leverage the VAE encoder $\mathcal{E}$ from pretrained LDM to obtain $z_i(T)$ for each specific real sample $x_i$. We first extract the low-dimensional latent representation by feeding the training image into VAE encoder: $z_i(0) \gets \mathcal{E}(x_i)$. We implement additional measures to enhance client privacy. First, we interpolate $z_i(0)$ with another latent representation, $z_{i'}(0)$, from the same class, thereby reducing the risk of exact sample reconstruction. Second, we add $T$-steps of random noise to obtain $z_i(T)$, which corresponds to the maximum noise intensity in LDM. A comprehensive privacy analysis is provided in Section \ref{sec:privacy} and \ref{sec:visualization}. The overall process can be formalized as 
\begin{equation}
z_i(T) \gets \delta(T, \gamma z_i(0) + (1-\gamma) z_{i'}(0)), s.t., i \neq i', y_i = y_{i'},
\end{equation}
where $\gamma \sim \mathcal{N}(0.5, 0.1^2)$ and clipped to $[0,1]$. After the computation, we store $z_i(T)$ and its corresponding ground truth label $y_i$ for all training images in the $k^{th}$ client as the instance-level personalization. We emphasize that this level of personalization does not require any additional optimization, making the process computationally efficient.

\begin{table*}[ht]
    \renewcommand\arraystretch{1.1}
    \caption{Evaluation results of different methods on three OSFL benchmarks with feature space heterogeneity. We report the mean{\scriptsize ±std} classification accuracy from 3 runs with different seeds. The best and second-best results are marked with \textbf{bold} and \underline{underline}, respectively.}
    \vspace{-5pt}
    \centering
    \footnotesize
    \setlength{\tabcolsep}{2.15mm}
    \begin{tabular}{c|c||cc|cccc||ccc}
        \toprule
        \multicolumn{2}{c||}{Dataset} & FedAvg & Central (\textit{oracle}) & FedD3 & DENSE & FedDEO & FGL & \textbf{FedBiP-S} &\textbf{ FedBiP-M} & \textbf{FedBiP-L}\\
        \hline
         \hline
         \multirow{7}*{\makecell[c]{Domain\\Net}} & C & \makecell[c]{73.12 {\tiny ±1.54}} & \makecell[c]{73.63 {\tiny ±0.91}} & \makecell[c]{61.21 {\tiny ±1.46}} & \makecell[c]{63.84 {\tiny ±2.51}} & \makecell[c]{72.33 {\tiny ±1.26}} & \makecell[c]{67.71 {\tiny ±3.15}} & \makecell[c]{68.07 {\tiny ±0.96}} & \makecell[c]{\underline{74.01} {\tiny ±1.67}} & \makecell[c]{\textbf{77.52} {\tiny ±0.67}} \\
         ~ & I & \makecell[c]{59.85 {\tiny ±1.51}} & \makecell[c]{61.76 {\tiny ±0.94}} & \makecell[c]{50.39 {\tiny ±1.64}} & \makecell[c]{52.87 {\tiny ±0.38}} & \makecell[c]{57.39 {\tiny ±0.84}} & \makecell[c]{\underline{59.83} {\tiny ±1.55}} & \makecell[c]{54.06 {\tiny ±2.56}} & \makecell[c]{58.42 {\tiny ±2.05}} & \makecell[c]{\textbf{60.94} {\tiny ±2.08}} \\
         ~ & P & \makecell[c]{63.77 {\tiny ±1.12}} & \makecell[c]{69.18 {\tiny ±1.74}} & \makecell[c]{60.50 {\tiny ±1.09}} & \makecell[c]{62.07 {\tiny ±0.97}} & \makecell[c]{63.17 {\tiny ±1.05}} & \makecell[c]{\textbf{68.56} {\tiny ±2.51}} & \makecell[c]{58.24 {\tiny ±0.22}} & \makecell[c]{63.01 {\tiny ±2.25}} & \makecell[c]{\underline{65.20} {\tiny ±0.78}} \\
         ~ & Q & \makecell[c]{16.26 {\tiny ±2.60}} & \makecell[c]{72.83 {\tiny ±0.82}} & \makecell[c]{28.25 {\tiny ±3.11}} & \makecell[c]{29.92 {\tiny ±1.62}} & \makecell[c]{37.86 {\tiny ±2.47}}  & \makecell[c]{19.83 {\tiny ±2.99}} & \makecell[c]{\underline{51.09} {\tiny ±2.05}} & \makecell[c]{49.64 {\tiny ±5.05}} & \makecell[c]{\textbf{51.85} {\tiny ±3.24}} \\
         ~ & R & \makecell[c]{87.90 {\tiny ±0.09}} & \makecell[c]{87.86 {\tiny ±0.24}} & \makecell[c]{79.15 {\tiny ±1.44}} & \makecell[c]{81.69 {\tiny ±1.14}} & \makecell[c]{81.51 {\tiny ±1.03}}  & \makecell[c]{\textbf{87.09} {\tiny ±0.88}} & \makecell[c]{80.44 {\tiny ±1.38}} & \makecell[c]{82.20 {\tiny ±0.67}} & \makecell[c]{\underline{83.16} {\tiny ±0.60}} \\
         ~ & S & \makecell[c]{68.07} {\tiny ±4.67} & \makecell[c]{75.28 {\tiny ±0.96}} & \makecell[c]{58.07 {\tiny ±1.35}} & \makecell[c]{59.20 {\tiny ±2.12}} & \makecell[c]{{62.86} {\tiny ±1.61}} & \makecell[c]{\underline{67.15} {\tiny ±3.97}} & \makecell[c]{57.17 {\tiny ±1.59}} & \makecell[c]{61.92 {\tiny ±1.35}} & \makecell[c]{\textbf{68.24} {\tiny ±0.78}} \\
         \cline{2-11}         
         ~ & \textbf{Avg} & \makecell[c]{61.49 {\tiny ±0.58}} & \makecell[c]{73.42 {\tiny ±0.53}} & \makecell[c]{56.26 {\tiny ±0.74}} & \makecell[c]{58.26 {\tiny ±1.33}} & \makecell[c]{62.52 {\tiny ±1.56}} & \makecell[c]{61.69 {\tiny ±1.56}} & \makecell[c]{61.51 {\tiny ±0.62}} & \makecell[c]{\underline{64.86} {\tiny ±0.49}} & \makecell[c]{\textbf{67.82} {\tiny ±0.56}} \\
         \hline
         \hline
         \multirow{5}*{\makecell[c]{PACS}} & {A} & \makecell[c]{52.68 {\tiny ±3.22}} & \makecell[c]{53.06 {\tiny ±0.53}} & \makecell[c]{42.42 {\tiny ±1.81}} & \makecell[c]{44.64 {\tiny ±0.14}} & \makecell[c]{{49.89} {\tiny ±0.91}} & \makecell[c]{\textbf{55.04} {\tiny ±1.79}} & \makecell[c]{43.01 {\tiny ±1.80}} & \makecell[c]{50.15 {\tiny ±1.86}} & \makecell[c]{\underline{53.26} {\tiny ±2.54}} \\
         ~ & {C} & \makecell[c]{68.27 {\tiny ±4.22}} & \makecell[c]{71.43 {\tiny ±1.61}} & \makecell[c]{60.47 {\tiny ±2.46}} & \makecell[c]{63.10 {\tiny ±1.47}} & \makecell[c]{{68.31} {\tiny ±1.41}} & \makecell[c]{\underline{69.94} {\tiny ±1.43}} & \makecell[c]{64.58 {\tiny ±3.23}} & \makecell[c]{67.71 {\tiny ±0.93}} & \makecell[c]{\textbf{70.90} {\tiny ±2.97}} \\
         ~ & {P} & \makecell[c]{86.31 {\tiny ±1.03}} & \makecell[c]{81.55 {\tiny ±6.16}} & \makecell[c]{72.08 {\tiny ±2.25}} & \makecell[c]{74.70 {\tiny ±0.81}} & \makecell[c]{{71.96} {\tiny ±0.56}} & \makecell[c]{\textbf{76.47} {\tiny ±0.68}} & \makecell[c]{70.24 {\tiny ±2.73}} & \makecell[c]{73.07 {\tiny ±1.80}} & \makecell[c]{\underline{74.85} {\tiny ±1.36}} \\
         ~ & {S} & \makecell[c]{31.25 {\tiny ±9.94}} & \makecell[c]{63.24 {\tiny ±3.35}} & \makecell[c]{30.40 {\tiny ±1.99}} & \makecell[c]{31.40 {\tiny ±2.06}} & \makecell[c]{{48.95} {\tiny ±1.34}} & \makecell[c]{41.82 {\tiny ±6.26}} & \makecell[c]{48.66 {\tiny ±4.26}} & \makecell[c]{\underline{50.30} {\tiny ±2.20}} & \makecell[c]{\textbf{51.70} {\tiny ±1.69}} \\
         \cline{2-11}         
         ~ & \textbf{Avg} & \makecell[c]{59.63 {\tiny ±3.13}} & \makecell[c]{67.32 {\tiny ±2.36}} & \makecell[c]{51.34 {\tiny ±2.51}} & \makecell[c]{53.46 {\tiny ±1.62}} & \makecell[c]{{59.78} {\tiny ±1.07}} & \makecell[c]{\underline{60.82} {\tiny ±1.90}} & \makecell[c]{56.62 {\tiny ±1.23}} & \makecell[c]{{60.30} {\tiny ±0.42}} & \makecell[c]{\textbf{62.67} {\tiny ±0.45}} \\
         \hline
         \hline
         \multirow{5}*{\makecell[c]{Office\\Home}} & {A} & \makecell[c]{54.48 {\tiny ±1.60}} & \makecell[c]{58.68 {\tiny ±1.72}} & \makecell[c]{50.71 {\tiny ±1.30}} & \makecell[c]{\underline{52.37} {\tiny ±0.96}} &  \makecell[c]{{49.37} {\tiny ±2.06}}  & \makecell[c]{48.48 {\tiny ±3.18}} & \makecell[c]{39.80 {\tiny ±0.88}} & \makecell[c]{45.06 {\tiny ±0.75}} & \makecell[c]{\textbf{55.41} {\tiny ±0.55}} \\
         ~ & {C} & \makecell[c]{{47.63} {\tiny ±1.08}} & \makecell[c]{51.09 {\tiny ±1.17}} & \makecell[c]{44.06 {\tiny ±0.86}} & \makecell[c]{\underline{46.24} {\tiny ±1.74}} &  \makecell[c]{{42.92} {\tiny ±0.81}}  & \makecell[c]{36.58 {\tiny ±2.36}} & \makecell[c]{36.79 {\tiny ±1.15}} & \makecell[c]{40.86 {\tiny ±0.80}} & \makecell[c]{\textbf{48.62} {\tiny ±0.42}} \\
         ~ & {P} & \makecell[c]{{73.94} {\tiny ±1.27}} & \makecell[c]{77.79 {\tiny ±0.83}} & \makecell[c]{71.09 {\tiny ±1.69}} & \makecell[c]{73.76 {\tiny ±2.07}} &  \makecell[c]{\underline{73.81} {\tiny ±0.46}}  & \makecell[c]{{59.38} {\tiny ±0.66}} & \makecell[c]{69.20 {\tiny ±1.17}} & \makecell[c]{73.23 {\tiny ±0.69}} & \makecell[c]{\textbf{76.63} {\tiny ±0.20}} \\
         ~ & {R} & \makecell[c]{{63.94} {\tiny ±0.56}} & \makecell[c]{69.97 {\tiny ±0.63}} & \makecell[c]{60.25 {\tiny ±0.88}} & \makecell[c]{61.86 {\tiny ±1.45}} &  \makecell[c]{{61.77} {\tiny ±0.51}}  & \makecell[c]{\underline{62.08} {\tiny ±2.37}} & \makecell[c]{56.57 {\tiny ±1.01}} & \makecell[c]{61.94 {\tiny ±1.32}} & \makecell[c]{\textbf{65.43} {\tiny ±0.96}} \\
         \cline{2-11}         
         ~ & \textbf{Avg} & \makecell[c]{{60.00} {\tiny ±0.88}} & \makecell[c]{64.38 {\tiny ±1.06}} & \makecell[c]{56.52 {\tiny ±1.07}} & \makecell[c]{\underline{58.55} {\tiny ±1.35}} &  \makecell[c]{{56.96} {\tiny ±1.71}}  & \makecell[c]{{51.63} {\tiny ±1.71}} & \makecell[c]{50.59 {\tiny ±0.70}} & \makecell[c]{{55.27} {\tiny ±0.73}} & \makecell[c]{\textbf{61.52} {\tiny ±0.39}} \\
        \bottomrule
    \end{tabular}
    \vspace{-8pt}
    \label{tab:mainresults}
\end{table*}

\subsubsection{Concept-level Personalization}
Solely applying instance-level personalization results in reduced diversity in image generation. To mitigate this limitation, we enhance personalization by incorporating domain and category concepts into the LDM generation process. Specifically, "domain" denotes the feature distribution within a client's local dataset, such as an image style in the DomainNet dataset. To avoid the costly finetuning of the LDM weights $\theta$, we finetune only the textual guidance. Specifically, we randomly initialize the domain concept vector $S\in\mathbb{R}^{n_s \times d_w}$ and category concept vector $V\in\mathbb{R}^{C \times n_v \times d_w}$, where $n_s$ and $n_v$ are the number of tokens for domain concept and category concept, respectively, and $d_w$ is the token embedding dimension of the textual conditioning model $\tau_{\theta}$. Subsequently, specific tokens in the textual template $P$ are substituted with the concept vectors $S$ and $V_y$ corresponding to a specific category $y$. For instance, this could result in textual prompts like  \textit{''A $[S]$ style of a $[V_y]$''} for DomainNet dataset. Following this, $\tau_{\theta}$ encodes these modified prompts, transforming the textual embeddings into intermediate representation for the denoising UNet $\epsilon_{\theta}$. 


To jointly optimize both concept vectors $S$ and $V_y$, we adopt the following objective function:
\begin{equation}
L_g = \mathbb{E}_{\mathcal{E}(x(0)),y,\epsilon \sim \mathcal{N}(0,1), t}\left[||\epsilon - \epsilon_{\theta}(z(t), t, \tau_{\theta}(S, V_y))||_2^2\right],
\label{eq:main}
\end{equation}
where timestep $t$ is sampled from $\text{Uniform}(\{1,...,T\})$. 

After the local optimization of each client, the latent vectors $\{z_i(T), y_i\}_{i=1}^{N^k}$, along with the optimized concept vectors $S, V$, are uploaded to the central server. To further increase the generation diversity, we introduce a small perturbation to the domain concept vector $S$. Specifically, we define $\hat{S} = S + \eta$ with $\eta \sim \mathcal{N}(0,\sigma_\eta)$, where $\sigma_\eta$ controls the perturbation intensity. The central server then integrates these vectors into the same pretrained LDM and generates synthetic images with
\begin{equation}
\tilde{x_i} = \mathcal{D}(\epsilon_{\theta}(z_i(T), T, \tau_{\theta}(\hat{S}, V_{y_i}))).
\end{equation}
The data sample $(\tilde{x_i}, y_i)$ is appended to the synthetic set $D_{syn}$. It is crucial to note that \texttt{FedBiP} performs image generation asynchronously, eliminating the need to wait for all clients to complete their local processes. Once the server receives the vectors uploaded from all clients and completes the image generation, we proceed to optimize the target classification model $\phi$ with the objective:
\begin{equation}
\label{eq:cls}
L_{cls} = L_{CE}(\phi(\tilde{x}), y).
\end{equation}

\section{Experiments and Analyses}

We conduct extensive empirical analyses to investigate the proposed method. Firstly, we compare \texttt{FedBiP} with other baseline methods on three One-Shot Federated Learning (OSFL) benchmarks with feature space heterogeneity. Next, we evaluate \texttt{FedBiP} using a medical dataset and a satellite image dataset adapted for OSFL setting with label space heterogeneity, illustrating its effectiveness under challenging real-world scenarios. Finally, we perform an ablation study on \texttt{FedBiP} and further analyze its promising privacy-preserving capability.

\subsection{Benchmark Experiments}
\textbf{Datasets Description:} We adapt three common image classification benchmarks with feature distribution shift for our OSFL setting: (1) \textit{DomainNet} \citep{peng2019moment}, which contains six domains: Clipart (C), Infograph (I), Painting (P), Quickdraw (Q), Real (R), and Sketch (S). We select 10 categories following \citep{zhang2023federated}. (2) \textit{PACS} \citep{li2017deeper}, which includes images that belong to 7 classes from four domains: Art (A), Cartoon (C), Photo (P), and Sketch (S). (3) \textit{OfficeHome} \citep{venkateswara2017deep} comprises images of daily objects from four domains: Art (A), Clipart (C), Product (P), and Real (R). Each client is assigned a specific domain. To simulate local data scarcity described in previous sections, we adopt 16-shot per class (8-shot for OfficeHome) for each client, following \citep{li2021fedbn, chen2023fraug}.

\textbf{Baseline Methods:} We compare \texttt{FedBiP} with several baseline methods, including \textit{FedAvg} and \textit{Central}, i.e., aggregating the training data from all clients. We note that \textit{Central} is an oracle method as it infringes on privacy requirements, while \textit{FedAvg} requires multi-round communication and is not applicable to OSFL. Besides, we validate concurrent generation-based methods for OSFL: (1) \textit{FedD3} \citep{song2023federated}, where distilled instances from the clients are uploaded. (2) \textit{DENSE} \citep{zhang2022dense}, where client local models are uploaded and distilled into one model using synthetic images. (3) \textit{FedDEO} \citep{yang2024feddeo}, where the optimized category descriptions are uploaded and guide pretrained diffusion models. (4) \textit{FGL} \citep{zhang2023federated}, where captions of client local images, extracted by BLIP-2 \citep{li2023blip}, are uploaded and guide pretrained LDM. 

\textbf{Implementation Details:} We adopt the HuggingFace open-sourced "CompVis/stable-diffusion-v1-4" as the pretrained Latent Diffusion Model, and use ResNet-18 pretrained on ImageNet \citep{deng2009imagenet} as the initialization for the classification model. We investigate three variants of \texttt{FedBiP}, namely "S", "M", and "L", which corresponds to generating $2 \times$, $5 \times$, $10 \times$ the number of images in the original client local dataset, respectively. Note that synthesizing more images does not affect the client's local optimization costs. We optimize the concept vectors for 50 epochs at each client. For \textit{FGL}, 3500 samples per class per domain are generated. For \textit{FedDEO}, the total number of synthetic images is identical to \texttt{FedBiP-L} for a fair comparison. Further details about training hyperparameters are provided in Appendix.

\begin{table*}[t]
    \vspace{-8pt}
    \renewcommand\arraystretch{1.1}
    \caption{Evaluation results of different methods on real-world medical and satellite OSFL benchmarks with varying levels of label space heterogeneity. The best results are marked with \textbf{bold}.}
    \vspace{-5pt}
    \centering
    \footnotesize
    \setlength{\tabcolsep}{1.95mm}
    \begin{tabular}{c|c||cc|cccc||ccc}
        \toprule
        Dataset & Split & FedAvg & Central (\textit{oracle}) & FedD3 & DENSE & FedDEO & FGL & \textbf{FedBiP-S} & \textbf{FedBiP-M} & \textbf{FedBiP-L} \\
        \hline
        \hline
        \multirow{3}*{UCM} & IID & 63.82 {\tiny ±0.67} & 68.44 {\tiny ±0.52} &  59.37 {\tiny ±1.24}  & 64.08 {\tiny ±0.95}  & 63.15 {\tiny ±0.86} & 52.65 {\tiny ±1.74}  & 61.58 {\tiny ±0.76}  & 63.74 {\tiny ±0.47}  & \textbf{65.59} {\tiny ±1.01} \\
         ~ & $Dir_{0.5}$ & 62.96 {\tiny ±1.41} & 68.44 {\tiny ±0.52} & 56.86 {\tiny ±0.81}  & 61.41 {\tiny ±1.51}  & 61.04 {\tiny ±0.34}  & 52.65 {\tiny ±1.74}  & 61.02 {\tiny ±1.03}  & 62.37 {\tiny ±0.84}  & \textbf{64.41} {\tiny ±0.88} \\
         ~ & $Dir_{0.01}$ & 57.47 {\tiny ±1.76} & 68.44 {\tiny ±0.52} & 50.24 {\tiny ±0.49}  & 54.16 {\tiny ±0.77}  & 55.81 {\tiny ±1.05}  &  52.65 {\tiny ±1.74}  & 54.48 {\tiny ±1.24}  & 56.19 {\tiny ±0.65}  & \textbf{59.84} {\tiny ±0.47} \\
        \hline
        \hline
        \multirow{3}*{\makecell[c]{Derma\\MNIST}}  & IID & 53.47 {\tiny ±1.49} & 60.08 {\tiny ±0.98} & 50.26 {\tiny ±0.67}  & 52.91 {\tiny ±0.34}  & 54.29 {\tiny ±1.12} & 40.82 {\tiny ±2.56}  & 53.84 {\tiny ±1.52}  & 54.91 {\tiny ±0.71}  & \textbf{56.10} {\tiny ±1.34}   \\
         ~ & $Dir_{0.5}$  & 51.98 {\tiny ±0.52} & 60.08 {\tiny ±0.98} & 49.52 {\tiny ±1.46}  & 50.83 {\tiny ±0.61}  & 52.61 {\tiny ±0.84} & 40.82 {\tiny ±2.56}  & 51.47 {\tiny ±1.32}  & 53.26 {\tiny ±0.84}  & \textbf{55.03} {\tiny ±1.02} \\
         ~ & $Dir_{0.01}$ & 43.99 {\tiny ±2.07} & 60.08 {\tiny ±0.98} & 40.25 {\tiny ±1.91}  & 41.08 {\tiny ±2.30}  & 42.14 {\tiny ±0.96} & 40.82 {\tiny ±2.56}  & 45.32 {\tiny ±0.91}  & 46.71 {\tiny ±1.31}  & \textbf{48.15} {\tiny ±1.67} \\
        \bottomrule
    \end{tabular}
    \label{tab:medresults}
    \vspace{-10pt}
\end{table*}

\textbf{Results and Analyses:} We report the validation results in Table \ref{tab:mainresults}, where we observe \texttt{FedBiP-L} outperforms all baseline methods in average performance, indicating an average performance improvement of up to $5.96\%$. Notably, \texttt{FedBiP-S} achieves comparable performance to \textit{FGL} by generating only 16 images for DomainNet per class and domain, while \textit{FGL} requires 3500 images. This further highlights the efficiency of our proposed method. Additionally, \texttt{FedBiP} excels in challenging domains, such as Quickdraw (Q) of DomainNet and Sketch (S) of PACS, showcasing its effectiveness in generating images that are rare in the Latent Diffusion Model (LDM) pretraining dataset. However, \texttt{FedBiP} slightly underperforms in certain domains, e.g., Real (R) in DomainNet. We attribute this to the overlap between these domains and the LDM pretraining dataset, where adapting LDM with the client local datasets reduces its generation diversity. Nevertheless, \texttt{FedBiP} narrows the gap between the generation-based methods and oracle \texttt{Central} method.

\subsection{Validation on Medical and Satellite Image Datasets}
To illustrate the effectiveness of \texttt{FedBiP} on challenging real-world applications, we adopt a medical dataset, \textit{DermaMNIST} \citep{medmnistv2}, comprising dermatoscopic images of 7 types of skin lesion, and a satellite image dataset, UC Merced Land Use Dataset (\textit{UCM}) \citep{yang2010bag}, which includes satellite images representing 21 different land use categories. We assume there are 5 research institutions (clients) participating in the collaborative training. To construct local datasets for each client in OSFL, we employ the Dirichlet distribution $Dir_\beta$ to model label space heterogeneity, in which a smaller $\beta$ indicates higher data heterogeneity. Following \citep{zhou2022learning}, we use the textual template "\textit{A dermatoscopic image of a [CLS], a type of pigmented skin lesions.}" and "\textit{A centered satellite photo of [CLS].}" for DermaMNIST and UCM, respectively.

In Table \ref{tab:medresults}, we report the validation results of different methods on real-world OSFL benchmarks with varying levels of label space heterogeneity. We observe that \texttt{FedBiP-L} consistently outperforms all baseline methods across all settings, with an average performance increase of up to $4.16\%$ over \textit{FedAvg}. Furthermore, we notice that the most lightweight version, \texttt{FedBiP-S}, surpasses the method with pretrained LDM, \textit{FGL}, by a substantial margin. This demonstrates the importance of our LDM personalization schema, particularly in scenarios involving significant feature distribution shifts compared to the pretraining dataset of LDM.

\subsection{Ablation Study}
To illustrate the importance of different \texttt{FedBiP} components, we conduct an ablation study on three OSFL benchmark datasets. The results are shown in Table \ref{tab:ablation}. First, we observe that simply prompting LDM with \textit{''A [STY] style of a [CLS]''} and synthesizing images at central server is ineffective. Next, we notice that optimizing only the category concept vector $V_c$ leads to only minimal performance improvements. We hypothesize that this is because the categories in these benchmarks are general objects, such as "person" or "clock", which are already well-captured by LDM during pretraining. In contrast, optimizing the domain concept vector $S$ produces visible performance gain. This can be attributed to the mismatch between the textual representation of domain concepts and LDM's pretraining. For example, as described in Motivation section (Figure \ref{fig:motivation}), "Quickdraw" in DomainNet encompasses images characterized by very simple lines, while LDM tends to generate images with finer details. Furthermore, applying instance-level personalization with $z(T)$ yields a performance boost, highlighting the importance of fine-grained personalization in improving LDM. Finally, combining both levels of personalization further improves the results, which demonstrates their complementarity.

\begin{table}[b]
    \centering
    \vspace{-10pt}
    \caption{Ablation study for different components of \texttt{FedBiP}.}
    \footnotesize
    \label{tab:ablation}    
    \vspace{-5.5pt}
    \setlength{\tabcolsep}{2.4mm}{
        \begin{tabular}{ccc|ccc}
    \toprule
    \textbf{Instance} & \multicolumn{2}{c|}{\qquad \textbf{Concept}} & \multirow{2}*{\makecell[c]{Domain\\Net}} & \multirow{2}*{PACS} & \multirow{2}*{\makecell[c]{Office\\Home}} \\
    $z(T)$ & $\hat{S}$ & $V_c$ \\
    \hline
    \multicolumn{3}{c|}{FedAvg (\textit{multi-round})} & \makecell[c]{61.49} & \makecell[c]{59.63}  & \makecell[c]{60.00}  \\ 
    \hline
     & & & \makecell[c]{60.22} & \makecell[c]{58.90}  & \makecell[c]{53.23} \\ 
    \hline
     & & \checkmark & \makecell[c]{61.71} & \makecell[c]{59.15} & \makecell[c]{55.81} \\
     \hline 
     & \checkmark & & \makecell[c]{63.96} & \makecell[c]{60.08} & \makecell[c]{56.32} \\
     \hline
    \checkmark & & & \makecell[c]{66.08} & \makecell[c]{61.83} & \makecell[c]{59.35}\\ 
    \hline
    \checkmark & \makecell[c]{\checkmark (no perturb.)} & \checkmark & \makecell[c]{67.09} & \makecell[c]{62.78} & \makecell[c]{60.84} \\
    \hline
    \checkmark & \checkmark & \checkmark & \makecell[c]{67.82} & \makecell[c]{62.67} & \makecell[c]{61.52} \\
    \bottomrule
    \end{tabular}}
\end{table}

\subsection{Scalability Analysis of \texttt{FedBiP}}
To show the scalability of \texttt{FedBiP} under various application scenarios, we validate \texttt{FedBiP} in systems with varying client numbers and analyze the effects of synthetic image quantity. 

\textbf{Varying Number of Clients:} We split each domain of the DomainNet dataset into 5 subsets, ensuring that each subset contains 16 samples per category to simulate the local data scarcity described in previous sections. Each subset is then assigned to a specific client. In our experiments, we select 1 to 5 clients from each domain, resulting in a total of 6 to 30 clients participating in federated learning. 

\begin{figure}[t]
    \centering 
    \vspace{-10pt}
    \includegraphics[width=0.7\linewidth]{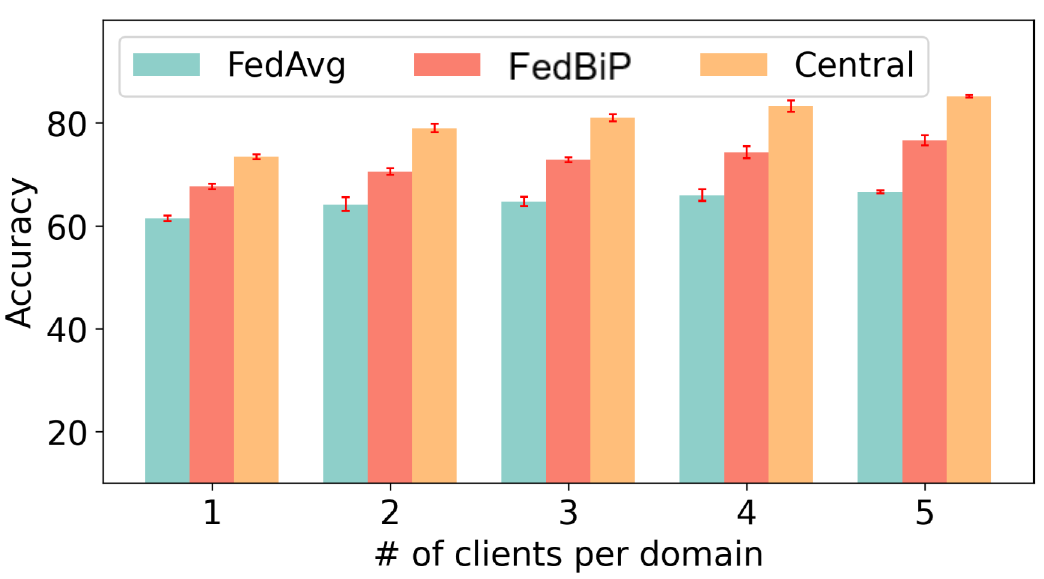}
    \vspace{-10pt}
    \caption{Validation results with varying client numbers on DomainNet.}
    \vspace{-15pt}
    \label{fig:multiclient}
\end{figure}

The validation results are presented in Figure \ref{fig:multiclient}. We observe that the performance of the baseline method \textit{FedAvg} remains unchanged with the addition of more clients to FL. In contrast, the validation performance of \texttt{FedBiP} consistently increases, narrowing the gap between distributed optimization and \textit{Central} optimization. Furthermore, \texttt{FedBiP} outperforms \textit{FedAvg} by $9.51\%$ when the largest number of clients join FL, further indicating its scalability for real-world complex federated systems with more clients.

\begin{figure}[b]
    \vspace{-12pt}
    \centering
    \includegraphics[width=\linewidth]{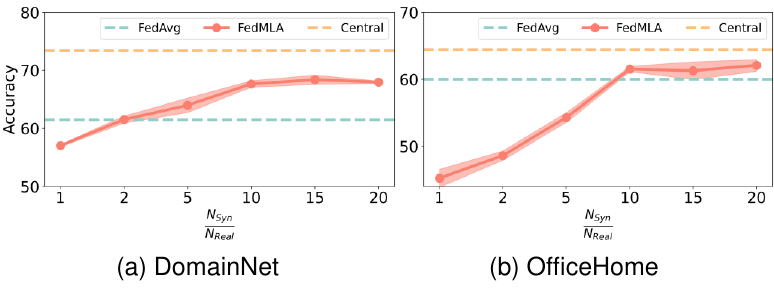}
    \vspace{-15pt}
    \caption{Validation results with synthesizing different numbers of images at central server.}
    \label{fig:varyimgnum}
\end{figure}

\textbf{Varying Number of Synthetic Images:}
We synthesize varying quantities of images for each category and domain, scaling from $1 \times$ to $20 \times$ the size of the original client local dataset. The results for the DomainNet and OfficeHome benchmarks are presented in Figure \ref{fig:varyimgnum}. Our analysis reveals that increasing the number of synthetic images enhances the performance of the target classification model, significantly outperforming the baseline method (\textit{FedAvg}) by up to $6.47\%$. Furthermore, we observe that synthesizing images at $10 \times$ the original dataset size emerges as the most effective approach, when considering the trade-off between generation time and final performance. This finding is consistent with the design principles of \texttt{FedMLA-L}.

\begin{figure*}[t]
\centering
\includegraphics[scale=0.475]{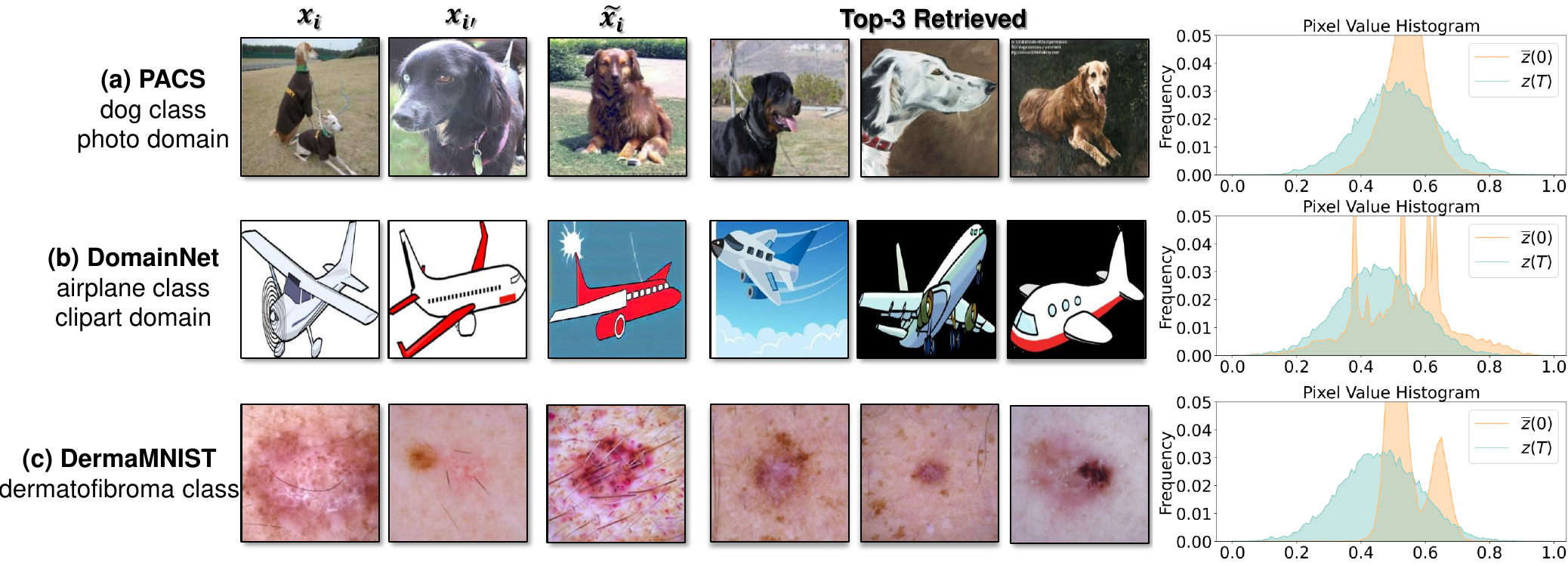} 
\vspace{-6pt}
\caption{\texttt{FedBiP} privacy analysis: (1) \textbf{Visual}: The reproduced images are notably dissimilar to the original images $x_i$ and $x_{i'}$. Besides, the retrieved images exhibit visual discrepancies compared to synthetic $\tilde{x}_i$. (2) \textbf{Statistical}: The pixel value histogram of $z(T)$ resembles a standard Gaussian distribution more closely compared to $\overline{z}(0)$, making it hard to extract private information from $z(T)$. }
\vspace{-5pt}
\label{fig:privacy} 
\end{figure*}

\subsection{Privacy Analysis}
\label{sec:privacy}
In this section, we present a comprehensive privacy analysis of \texttt{FedBiP}, encompassing both qualitative and quantitative evaluations, as illustrated in Figure \ref{fig:privacy}. 

\textbf{Visual discrepancy between synthetic and real images}: We visualize both synthetic image $\tilde{x}_i$, and its corresponding real images, i.e., $x_i$, $x_{i'}$. Besides, we use the pretrained ResNet-18 to extract the feature map of $\tilde{x}_i$ and retrieve the top-3 real images which indicate the largest cosine similarities in the feature space. We observe differences in both background (e.g., textual and color) and foreground (e.g., the exact object shape, position, and pose) between real and synthetic images. These visual discrepancies indicate that the synthetic images do not closely resemble any individual real images, thereby reducing the risk of revealing sensitive information about the original client data.

\textbf{Pixel Value Histogram Analysis}: To further analyze \texttt{FedBiP} from a statistical perspective, we provide histograms of both $\overline{z}(0)$ (the interpolated latent vectors of input images) and the corresponding $z(T)$ ($\overline{z}(0)$ with $T$-steps of random noise added). We observe that $z(T)$ closely resembles a standard Gaussian distribution, which contains less information about the original input images compared to $\overline{z}(0)$. This indicates that transmitting the noised $z(T)$ is more private than $\overline{z}(0)$, and would not significantly compromise privacy regulations. Additionally, we notice that $\overline{z}(0)$ could be further replaced with the average latent vectors of all samples from a specific class, i.e., categorical prototypes \citep{tan2022fedproto}. This substitution might further protect client privacy and is appropriate for applications with stringent privacy requirements. We leave this for future work.

\textbf{Membership Inference Attack (MIA) Analysis}: Finally, we analyze the resilience of \texttt{FedBiP} against MIA. Following \citep{yeom2018privacy, salem2018ml}, we compute the average loss and entropy of the final model on both training member and non-member data, and report the difference between the two averages. A smaller difference corresponds to better membership privacy preservation. From the MIA Analysis in Table \ref{tab:mia}, we can observe that \texttt{FedBiP} demonstrates superior resilience against MIA.

\begin{table}
\centering
\vspace{-5pt}
\renewcommand\arraystretch{1.1}
\footnotesize
\setlength\tabcolsep{8pt}
\caption{Membership Inference Attack (MIA) analysis on different benchmarks. A lower metric corresponds to better MIA privacy.}
\vspace{-5pt}
\begin{tabular}{c|c|cc}
\toprule
Dataset & MIA Metric & FedAvg & \textbf{FedBiP} \\
\hline
\multirow{2}*{\makecell[c]{DomainNet}}  & Entropy $\downarrow$  & 0.1311 & \makecell[c]{ 0.0186 \textbf{{\scriptsize $\downarrow$85.8\%}}}  \\ 
~ & Loss $\downarrow$  & 0.5976 & \makecell[c]{ 0.1611 \textbf{{\scriptsize $\downarrow$73.0\%}}}\\ 
\hline
\multirow{2}*{\makecell[c]{DermaMNIST}} & Entropy $\downarrow$ & 0.0897 &  \makecell[c]{0.0551 \textbf{{\scriptsize $\downarrow$38.6\%}}}  \\ 
~ & Loss $\downarrow$ & 0.5860 & \makecell[c]{0.4127 \textbf{{\scriptsize $\downarrow$29.6\%}}}\\ 
\hline
\multirow{2}*{\makecell[c]{PACS}} &  Entropy $\downarrow$  & 0.1635 &  \makecell[c]{0.0338 \textbf{{\scriptsize $\downarrow$79.3\%}}}  \\ 
~ & Loss $\downarrow$  & 0.4459 & \makecell[c]{0.1244 \textbf{{\scriptsize $\downarrow$72.1\%}}}\\ 
\bottomrule
\end{tabular}
\label{tab:mia}
\end{table}

\subsection{Visualization with Varying $\gamma$}
\label{sec:visualization}
\begin{figure}[b]
\vspace{-10pt}
\includegraphics[width=0.9\linewidth]{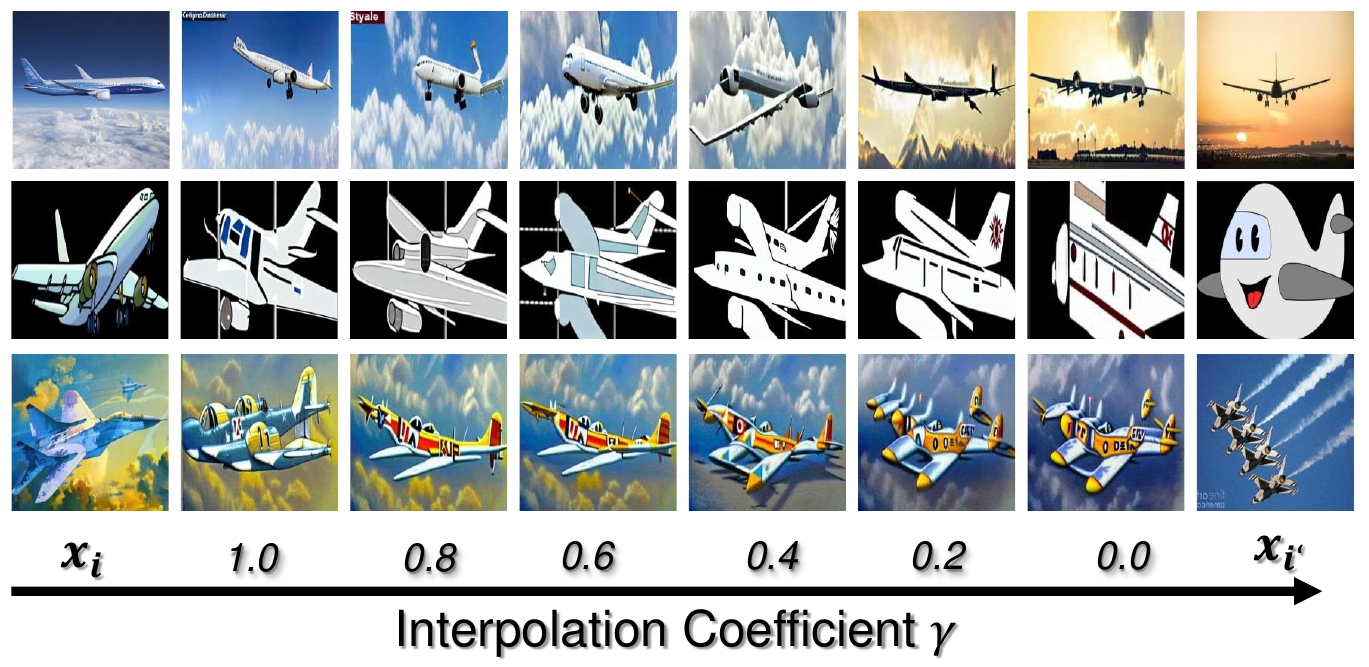} 
\vspace{-10pt}
\centering
\caption{Synthetic images generated with varying $\gamma$ for latent embedding interpolation.}
\vspace{-8pt}
\label{fig:interpolate} 
\end{figure}

In this section, we visualize the synthetic image $\tilde{x}_i$ using different interpolation coefficients $\gamma$ for DomainNet benchmark. Specifically, we compute the interpolated latent vector $\overline{z}_i(0)$ using $\gamma z_i(0) + (1-\gamma) z_{i'}(0)$. As shown in Figure \ref{fig:interpolate}, we observe that the synthetic images exhibit distinct visual characteristics compared to the real images, even when $\gamma$ is set to 0.0 or 1.0, corresponding to the direct use of latent vectors from the original images. We attribute these differences to the sampling process involved in the denoising phase of Latent Diffusion Model. Additionally, applying $\gamma$ values near 0.5 offers the most effective privacy protection. Most importantly, varying $\gamma$ produces diverse images, which enhances generation diversity and is beneficial for training the classification model. Therefore, we use a Gaussian distribution $\mathcal{N}(0.5, 0.1^2)$ to sample $\gamma$ in \texttt{FedBiP}.

\subsection{Visualization for Challenging Domains}
\begin{figure}[ht]
\centering
\includegraphics[width=\linewidth]{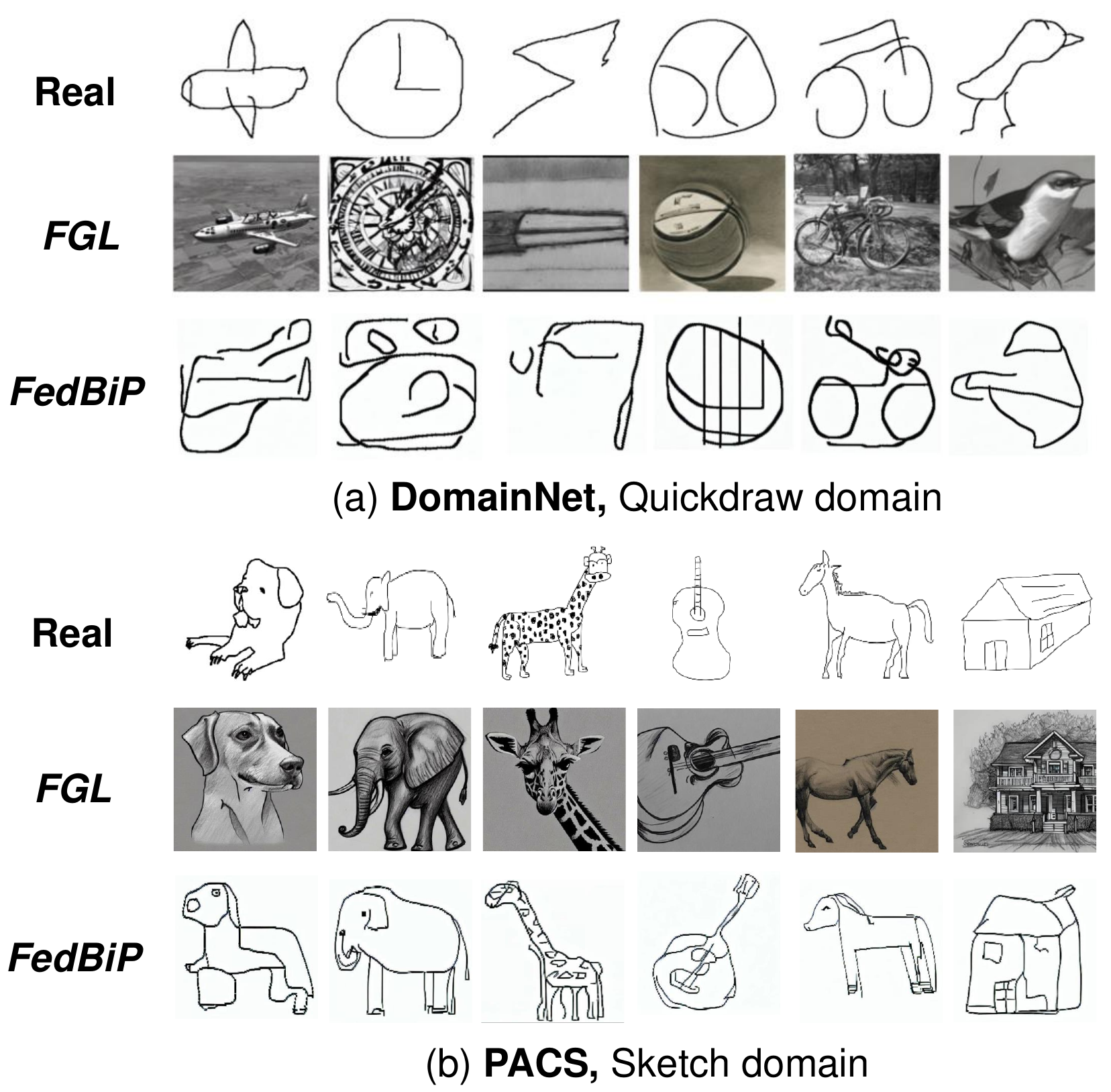} 
\vspace{-15pt}
\caption{Comparison of synthetic images from different algorithms for challenging domains.}
\label{fig:compare}
\vspace{-10pt}
\end{figure}

In this section, we present the synthetic images generated for the challenging domains, i.e., Quickdraw (DomainNet) and Sketch (PACS), as shown in Figure \ref{fig:compare}. Our observations indicate that \texttt{FedBiP} achieves superior generation quality by more accurately adhering to the original distribution of clients' local data compared to the diffusion-based method FGL \citep{zhang2023federated}. This visualization further highlights the effectiveness of our bi-level personalization approach.

\section{Conclusion}
In this work, we propose the first framework to address feature space heterogeneity in One-Shot Federated Learning (OSFL) using generative foundation models, specifically Latent Diffusion Model (LDM). The proposed method, named \texttt{FedBiP}, personalizes the pretrained LDM at both instance-level and concept-level. This design enables LDM to synthesize images that adhere to the local data distribution of each client, exhibiting significant deviations compared to its pretraining dataset. The experimental results indicate its effectiveness under OSFL systems with both feature and label space heterogeneity, surpassing the baseline and multiple concurrent methods. Additional experiments with medical or satellite images demonstrate its maturity for challenging real-world applications. Moreover, additional analysis highlights its promising scalability for more complex OSFL systems and privacy-preserving capability against different types of attacks.
{
    \small
    \bibliographystyle{ieeenat_fullname}
    \bibliography{main}

\begin{thebibliography}{54}
\providecommand{\natexlab}[1]{#1}
\providecommand{\url}[1]{\texttt{#1}}
\expandafter\ifx\csname urlstyle\endcsname\relax
  \providecommand{\doi}[1]{doi: #1}\else
  \providecommand{\doi}{doi: \begingroup \urlstyle{rm}\Url}\fi

\bibitem[Azizi et~al.(2023)Azizi, Kornblith, Saharia, Norouzi, and Fleet]{azizi2023synthetic}
Shekoofeh Azizi, Simon Kornblith, Chitwan Saharia, Mohammad Norouzi, and David~J Fleet.
\newblock Synthetic data from diffusion models improves imagenet classification.
\newblock \emph{arXiv preprint arXiv:2304.08466}, 2023.

\bibitem[Beitollahi et~al.(2024)Beitollahi, Bie, Hemati, Brunswic, Li, Chen, and Zhang]{beitollahi2024parametric}
Mahdi Beitollahi, Alex Bie, Sobhan Hemati, Leo~Maxime Brunswic, Xu Li, Xi Chen, and Guojun Zhang.
\newblock Parametric feature transfer: One-shot federated learning with foundation models.
\newblock \emph{arXiv preprint arXiv:2402.01862}, 2024.

\bibitem[Chen et~al.(2023)Chen, Frikha, Krompass, Gu, and Tresp]{chen2023fraug}
Haokun Chen, Ahmed Frikha, Denis Krompass, Jindong Gu, and Volker Tresp.
\newblock Fraug: Tackling federated learning with non-iid features via representation augmentation.
\newblock In \emph{Proceedings of the IEEE/CVF International Conference on Computer Vision}, pages 4849--4859, 2023.

\bibitem[Chen and Chao(2020)]{chen2020fedbe}
Hong-You Chen and Wei-Lun Chao.
\newblock Fedbe: Making bayesian model ensemble applicable to federated learning.
\newblock \emph{arXiv preprint arXiv:2009.01974}, 2020.

\bibitem[Croitoru et~al.(2023)Croitoru, Hondru, Ionescu, and Shah]{croitoru2023diffusion}
Florinel-Alin Croitoru, Vlad Hondru, Radu~Tudor Ionescu, and Mubarak Shah.
\newblock Diffusion models in vision: A survey.
\newblock \emph{IEEE Transactions on Pattern Analysis and Machine Intelligence}, 45\penalty0 (9):\penalty0 10850--10869, 2023.

\bibitem[Dai et~al.(2024)Dai, Zhang, Li, Liu, Yang, and Han]{dai2024enhancing}
Rong Dai, Yonggang Zhang, Ang Li, Tongliang Liu, Xun Yang, and Bo Han.
\newblock Enhancing one-shot federated learning through data and ensemble co-boosting.
\newblock \emph{arXiv preprint arXiv:2402.15070}, 2024.

\bibitem[Deng et~al.(2009)Deng, Dong, Socher, Li, Li, and Fei-Fei]{deng2009imagenet}
Jia Deng, Wei Dong, Richard Socher, Li-Jia Li, Kai Li, and Li Fei-Fei.
\newblock Imagenet: A large-scale hierarchical image database.
\newblock In \emph{2009 IEEE conference on computer vision and pattern recognition}, pages 248--255. Ieee, 2009.

\bibitem[Gong et~al.(2023)Gong, Danelljan, Sun, Mangas, and Van~Gool]{gong2023prompting}
Rui Gong, Martin Danelljan, Han Sun, Julio~Delgado Mangas, and Luc Van~Gool.
\newblock Prompting diffusion representations for cross-domain semantic segmentation.
\newblock \emph{arXiv preprint arXiv:2307.02138}, 2023.

\bibitem[Goodfellow et~al.(2020)Goodfellow, Pouget-Abadie, Mirza, Xu, Warde-Farley, Ozair, Courville, and Bengio]{goodfellow2020generative}
Ian Goodfellow, Jean Pouget-Abadie, Mehdi Mirza, Bing Xu, David Warde-Farley, Sherjil Ozair, Aaron Courville, and Yoshua Bengio.
\newblock Generative adversarial networks.
\newblock \emph{Communications of the ACM}, 63\penalty0 (11):\penalty0 139--144, 2020.

\bibitem[Guha et~al.(2019)Guha, Talwalkar, and Smith]{guha2019one}
Neel Guha, Ameet Talwalkar, and Virginia Smith.
\newblock One-shot federated learning.
\newblock \emph{arXiv preprint arXiv:1902.11175}, 2019.

\bibitem[Guo et~al.(2023)Guo, Yang, Rao, Wang, Qiao, Lin, and Dai]{guo2023animatediff}
Yuwei Guo, Ceyuan Yang, Anyi Rao, Yaohui Wang, Yu Qiao, Dahua Lin, and Bo Dai.
\newblock Animatediff: Animate your personalized text-to-image diffusion models without specific tuning.
\newblock \emph{arXiv preprint arXiv:2307.04725}, 2023.

\bibitem[Hasan et~al.(2024)Hasan, Zhang, Guo, Chen, and Poupart]{hasan2024calibrated}
Mohsin Hasan, Guojun Zhang, Kaiyang Guo, Xi Chen, and Pascal Poupart.
\newblock Calibrated one round federated learning with bayesian inference in the predictive space.
\newblock In \emph{Proceedings of the AAAI conference on artificial intelligence}, pages 12313--12321, 2024.

\bibitem[He et~al.(2016)He, Zhang, Ren, and Sun]{he2016deep}
Kaiming He, Xiangyu Zhang, Shaoqing Ren, and Jian Sun.
\newblock Deep residual learning for image recognition.
\newblock In \emph{Proceedings of the IEEE conference on computer vision and pattern recognition}, pages 770--778, 2016.

\bibitem[Ho et~al.(2020)Ho, Jain, and Abbeel]{ho2020denoising}
Jonathan Ho, Ajay Jain, and Pieter Abbeel.
\newblock Denoising diffusion probabilistic models.
\newblock \emph{Advances in neural information processing systems}, 33:\penalty0 6840--6851, 2020.

\bibitem[Humbert et~al.(2023)Humbert, Le~Bars, Bellet, and Arlot]{humbert2023one}
Pierre Humbert, Batiste Le~Bars, Aur{\'e}lien Bellet, and Sylvain Arlot.
\newblock One-shot federated conformal prediction.
\newblock In \emph{International Conference on Machine Learning}, pages 14153--14177. PMLR, 2023.

\bibitem[Kairouz et~al.(2021)Kairouz, McMahan, Avent, Bellet, Bennis, Bhagoji, Bonawitz, Charles, Cormode, Cummings, et~al.]{kairouz2021advances}
Peter Kairouz, H~Brendan McMahan, Brendan Avent, Aur{\'e}lien Bellet, Mehdi Bennis, Arjun~Nitin Bhagoji, Kallista Bonawitz, Zachary Charles, Graham Cormode, Rachel Cummings, et~al.
\newblock Advances and open problems in federated learning.
\newblock \emph{Foundations and trends{\textregistered} in machine learning}, 14\penalty0 (1--2):\penalty0 1--210, 2021.

\bibitem[Kang et~al.(2023)Kang, Chikontwe, Kim, Jin, Adeli, Pohl, and Park]{kang2023one}
Myeongkyun Kang, Philip Chikontwe, Soopil Kim, Kyong~Hwan Jin, Ehsan Adeli, Kilian~M Pohl, and Sang~Hyun Park.
\newblock One-shot federated learning on medical data using knowledge distillation with image synthesis and client model adaptation.
\newblock In \emph{International Conference on Medical Image Computing and Computer-Assisted Intervention}, pages 521--531. Springer, 2023.

\bibitem[Kasturi and Hota(2023)]{kasturi2023osgan}
Anirudh Kasturi and Chittaranjan Hota.
\newblock Osgan: One-shot distributed learning using generative adversarial networks.
\newblock \emph{The Journal of Supercomputing}, 79\penalty0 (12):\penalty0 13620--13640, 2023.

\bibitem[Kasturi et~al.(2020)Kasturi, Ellore, and Hota]{kasturi2020fusion}
Anirudh Kasturi, Anish~Reddy Ellore, and Chittaranjan Hota.
\newblock Fusion learning: A one shot federated learning.
\newblock In \emph{Computational Science--ICCS 2020: 20th International Conference, Amsterdam, The Netherlands, June 3--5, 2020, Proceedings, Part III 20}, pages 424--436. Springer, 2020.

\bibitem[Kawar et~al.(2023)Kawar, Zada, Lang, Tov, Chang, Dekel, Mosseri, and Irani]{kawar2023imagic}
Bahjat Kawar, Shiran Zada, Oran Lang, Omer Tov, Huiwen Chang, Tali Dekel, Inbar Mosseri, and Michal Irani.
\newblock Imagic: Text-based real image editing with diffusion models.
\newblock In \emph{Proceedings of the IEEE/CVF Conference on Computer Vision and Pattern Recognition}, pages 6007--6017, 2023.

\bibitem[Li et~al.(2017)Li, Yang, Song, and Hospedales]{li2017deeper}
Da Li, Yongxin Yang, Yi-Zhe Song, and Timothy~M Hospedales.
\newblock Deeper, broader and artier domain generalization.
\newblock In \emph{Proceedings of the IEEE international conference on computer vision}, pages 5542--5550, 2017.

\bibitem[Li et~al.(2023)Li, Li, Savarese, and Hoi]{li2023blip}
Junnan Li, Dongxu Li, Silvio Savarese, and Steven Hoi.
\newblock Blip-2: Bootstrapping language-image pre-training with frozen image encoders and large language models.
\newblock In \emph{International conference on machine learning}, pages 19730--19742. PMLR, 2023.

\bibitem[Li et~al.(2020)Li, He, and Song]{li2020practical}
Qinbin Li, Bingsheng He, and Dawn Song.
\newblock Practical one-shot federated learning for cross-silo setting.
\newblock \emph{arXiv preprint arXiv:2010.01017}, 2020.

\bibitem[Li et~al.(2021)Li, Jiang, Zhang, Kamp, and Dou]{li2021fedbn}
Xiaoxiao Li, Meirui Jiang, Xiaofei Zhang, Michael Kamp, and Qi Dou.
\newblock Fedbn: Federated learning on non-iid features via local batch normalization.
\newblock \emph{arXiv preprint arXiv:2102.07623}, 2021.

\bibitem[Li et~al.(2022)Li, Shao, Mao, Wang, and Zhang]{li2022federated}
Zijian Li, Jiawei Shao, Yuyi Mao, Jessie~Hui Wang, and Jun Zhang.
\newblock Federated learning with gan-based data synthesis for non-iid clients.
\newblock In \emph{International Workshop on Trustworthy Federated Learning}, pages 17--32. Springer, 2022.

\bibitem[Liu et~al.(2021)Liu, Chen, Qin, Dou, and Heng]{liu2021feddg}
Quande Liu, Cheng Chen, Jing Qin, Qi Dou, and Pheng-Ann Heng.
\newblock Feddg: Federated domain generalization on medical image segmentation via episodic learning in continuous frequency space.
\newblock In \emph{Proceedings of the IEEE/CVF conference on computer vision and pattern recognition}, pages 1013--1023, 2021.

\bibitem[Lyu et~al.(2020)Lyu, Yu, and Yang]{lyu2020threats}
Lingjuan Lyu, Han Yu, and Qiang Yang.
\newblock Threats to federated learning: A survey.
\newblock \emph{arXiv preprint arXiv:2003.02133}, 2020.

\bibitem[McInnes et~al.(2018)McInnes, Healy, and Melville]{mcinnes2018umap}
Leland McInnes, John Healy, and James Melville.
\newblock Umap: Uniform manifold approximation and projection for dimension reduction.
\newblock \emph{arXiv preprint arXiv:1802.03426}, 2018.

\bibitem[McMahan et~al.(2017)McMahan, Moore, Ramage, Hampson, and y~Arcas]{mcmahan2017communication}
Brendan McMahan, Eider Moore, Daniel Ramage, Seth Hampson, and Blaise~Aguera y Arcas.
\newblock Communication-efficient learning of deep networks from decentralized data.
\newblock In \emph{Artificial intelligence and statistics}, pages 1273--1282. PMLR, 2017.

\bibitem[Meng et~al.(2021)Meng, He, Song, Song, Wu, Zhu, and Ermon]{meng2021sdedit}
Chenlin Meng, Yutong He, Yang Song, Jiaming Song, Jiajun Wu, Jun-Yan Zhu, and Stefano Ermon.
\newblock Sdedit: Guided image synthesis and editing with stochastic differential equations.
\newblock \emph{arXiv preprint arXiv:2108.01073}, 2021.

\bibitem[Niemeijer et~al.(2024)Niemeijer, Schwonberg, Term{\"o}hlen, Schmidt, and Fingscheidt]{niemeijer2024generalization}
Joshua Niemeijer, Manuel Schwonberg, Jan-Aike Term{\"o}hlen, Nico~M Schmidt, and Tim Fingscheidt.
\newblock Generalization by adaptation: Diffusion-based domain extension for domain-generalized semantic segmentation.
\newblock In \emph{Proceedings of the IEEE/CVF Winter Conference on Applications of Computer Vision}, pages 2830--2840, 2024.

\bibitem[Peng et~al.(2019)Peng, Bai, Xia, Huang, Saenko, and Wang]{peng2019moment}
Xingchao Peng, Qinxun Bai, Xide Xia, Zijun Huang, Kate Saenko, and Bo Wang.
\newblock Moment matching for multi-source domain adaptation.
\newblock In \emph{Proceedings of the IEEE International Conference on Computer Vision}, pages 1406--1415, 2019.

\bibitem[Rombach et~al.(2022)Rombach, Blattmann, Lorenz, Esser, and Ommer]{rombach2022high}
Robin Rombach, Andreas Blattmann, Dominik Lorenz, Patrick Esser, and Bj{\"o}rn Ommer.
\newblock High-resolution image synthesis with latent diffusion models.
\newblock In \emph{Proceedings of the IEEE/CVF conference on computer vision and pattern recognition}, pages 10684--10695, 2022.

\bibitem[Salem et~al.(2018)Salem, Zhang, Humbert, Berrang, Fritz, and Backes]{salem2018ml}
Ahmed Salem, Yang Zhang, Mathias Humbert, Pascal Berrang, Mario Fritz, and Michael Backes.
\newblock Ml-leaks: Model and data independent membership inference attacks and defenses on machine learning models.
\newblock \emph{arXiv preprint arXiv:1806.01246}, 2018.

\bibitem[Sar{\i}y{\i}ld{\i}z et~al.(2023)Sar{\i}y{\i}ld{\i}z, Alahari, Larlus, and Kalantidis]{sariyildiz2023fake}
Mert~B{\"u}lent Sar{\i}y{\i}ld{\i}z, Karteek Alahari, Diane Larlus, and Yannis Kalantidis.
\newblock Fake it till you make it: Learning transferable representations from synthetic imagenet clones.
\newblock In \emph{Proceedings of the IEEE/CVF Conference on Computer Vision and Pattern Recognition}, pages 8011--8021, 2023.

\bibitem[Shin et~al.(2020)Shin, Hwang, Kim, Park, Bennis, and Kim]{shin2020xor}
MyungJae Shin, Chihoon Hwang, Joongheon Kim, Jihong Park, Mehdi Bennis, and Seong-Lyun Kim.
\newblock Xor mixup: Privacy-preserving data augmentation for one-shot federated learning.
\newblock \emph{arXiv preprint arXiv:2006.05148}, 2020.

\bibitem[So et~al.(2022)So, Hsieh, Arzani, Noghabi, Avestimehr, and Chandra]{so2022fedspace}
Jinhyun So, Kevin Hsieh, Behnaz Arzani, Shadi Noghabi, Salman Avestimehr, and Ranveer Chandra.
\newblock Fedspace: An efficient federated learning framework at satellites and ground stations.
\newblock \emph{arXiv preprint arXiv:2202.01267}, 2022.

\bibitem[Song et~al.(2023)Song, Liu, Chen, Festag, Trinitis, Schulz, and Knoll]{song2023federated}
Rui Song, Dai Liu, Dave~Zhenyu Chen, Andreas Festag, Carsten Trinitis, Martin Schulz, and Alois Knoll.
\newblock Federated learning via decentralized dataset distillation in resource-constrained edge environments.
\newblock In \emph{2023 International Joint Conference on Neural Networks (IJCNN)}, pages 1--10. IEEE, 2023.

\bibitem[Su et~al.(2023)Su, Li, and Xue]{su2023one}
Shangchao Su, Bin Li, and Xiangyang Xue.
\newblock One-shot federated learning without server-side training.
\newblock \emph{Neural Networks}, 164:\penalty0 203--215, 2023.

\bibitem[Tan et~al.(2022)Tan, Long, Liu, Zhou, Lu, Jiang, and Zhang]{tan2022fedproto}
Yue Tan, Guodong Long, Lu Liu, Tianyi Zhou, Qinghua Lu, Jing Jiang, and Chengqi Zhang.
\newblock Fedproto: Federated prototype learning across heterogeneous clients.
\newblock In \emph{Proceedings of the AAAI Conference on Artificial Intelligence}, pages 8432--8440, 2022.

\bibitem[Venkateswara et~al.(2017)Venkateswara, Eusebio, Chakraborty, and Panchanathan]{venkateswara2017deep}
Hemanth Venkateswara, Jose Eusebio, Shayok Chakraborty, and Sethuraman Panchanathan.
\newblock Deep hashing network for unsupervised domain adaptation.
\newblock In \emph{Proceedings of the IEEE Conference on Computer Vision and Pattern Recognition}, pages 5018--5027, 2017.

\bibitem[Yang et~al.(2023{\natexlab{a}})Yang, Shi, Wei, Liu, Zhao, Ke, Pfister, and Ni]{medmnistv2}
Jiancheng Yang, Rui Shi, Donglai Wei, Zequan Liu, Lin Zhao, Bilian Ke, Hanspeter Pfister, and Bingbing Ni.
\newblock Medmnist v2-a large-scale lightweight benchmark for 2d and 3d biomedical image classification.
\newblock \emph{Scientific Data}, 10\penalty0 (1):\penalty0 41, 2023{\natexlab{a}}.

\bibitem[Yang et~al.(2023{\natexlab{b}})Yang, Zhang, Song, Hong, Xu, Zhao, Zhang, Cui, and Yang]{yang2023diffusion}
Ling Yang, Zhilong Zhang, Yang Song, Shenda Hong, Runsheng Xu, Yue Zhao, Wentao Zhang, Bin Cui, and Ming-Hsuan Yang.
\newblock Diffusion models: A comprehensive survey of methods and applications.
\newblock \emph{ACM Computing Surveys}, 56\penalty0 (4):\penalty0 1--39, 2023{\natexlab{b}}.

\bibitem[Yang et~al.(2024{\natexlab{a}})Yang, Su, Li, and Xue]{yang2024exploring}
Mingzhao Yang, Shangchao Su, Bin Li, and Xiangyang Xue.
\newblock Exploring one-shot semi-supervised federated learning with pre-trained diffusion models.
\newblock In \emph{Proceedings of the AAAI Conference on Artificial Intelligence}, pages 16325--16333, 2024{\natexlab{a}}.

\bibitem[Yang et~al.(2024{\natexlab{b}})Yang, Su, Li, and Xue]{yang2024feddeo}
Mingzhao Yang, Shangchao Su, Bin Li, and Xiangyang Xue.
\newblock Feddeo: Description-enhanced one-shot federated learning with diffusion models.
\newblock In \emph{ACM Multimedia}, 2024{\natexlab{b}}.

\bibitem[Yang and Newsam(2010)]{yang2010bag}
Yi Yang and Shawn Newsam.
\newblock Bag-of-visual-words and spatial extensions for land-use classification.
\newblock In \emph{Proceedings of the 18th SIGSPATIAL international conference on advances in geographic information systems}, pages 270--279, 2010.

\bibitem[Yeom et~al.(2018)Yeom, Giacomelli, Fredrikson, and Jha]{yeom2018privacy}
Samuel Yeom, Irene Giacomelli, Matt Fredrikson, and Somesh Jha.
\newblock Privacy risk in machine learning: Analyzing the connection to overfitting.
\newblock In \emph{2018 IEEE 31st computer security foundations symposium (CSF)}, pages 268--282. IEEE, 2018.

\bibitem[Yuan et~al.(2023)Yuan, Zhang, Sun, Torr, and Zhao]{yuan2023real}
Jianhao Yuan, Jie Zhang, Shuyang Sun, Philip Torr, and Bo Zhao.
\newblock Real-fake: Effective training data synthesis through distribution matching.
\newblock \emph{arXiv preprint arXiv:2310.10402}, 2023.

\bibitem[Yurochkin et~al.(2019)Yurochkin, Agarwal, Ghosh, Greenewald, Hoang, and Khazaeni]{yurochkin2019bayesian}
Mikhail Yurochkin, Mayank Agarwal, Soumya Ghosh, Kristjan Greenewald, Nghia Hoang, and Yasaman Khazaeni.
\newblock Bayesian nonparametric federated learning of neural networks.
\newblock In \emph{International conference on machine learning}, pages 7252--7261. PMLR, 2019.

\bibitem[Zhang et~al.(2022)Zhang, Chen, Li, Lyu, Wu, Ding, Shen, and Wu]{zhang2022dense}
Jie Zhang, Chen Chen, Bo Li, Lingjuan Lyu, Shuang Wu, Shouhong Ding, Chunhua Shen, and Chao Wu.
\newblock Dense: Data-free one-shot federated learning.
\newblock \emph{Advances in Neural Information Processing Systems}, 35:\penalty0 21414--21428, 2022.

\bibitem[Zhang et~al.(2023)Zhang, Qi, and Zhao]{zhang2023federated}
Jie Zhang, Xiaohua Qi, and Bo Zhao.
\newblock Federated generative learning with foundation models.
\newblock \emph{arXiv preprint arXiv:2306.16064}, 2023.

\bibitem[Zhou et~al.(2022)Zhou, Yang, Loy, and Liu]{zhou2022learning}
Kaiyang Zhou, Jingkang Yang, Chen~Change Loy, and Ziwei Liu.
\newblock Learning to prompt for vision-language models.
\newblock \emph{International Journal of Computer Vision}, 130\penalty0 (9):\penalty0 2337--2348, 2022.

\bibitem[Zhou et~al.(2020)Zhou, Pu, Ma, Li, and Wu]{zhou2020distilled}
Yanlin Zhou, George Pu, Xiyao Ma, Xiaolin Li, and Dapeng Wu.
\newblock Distilled one-shot federated learning.
\newblock \emph{arXiv preprint arXiv:2009.07999}, 2020.

\bibitem[Zhu et~al.(2021)Zhu, Hong, and Zhou]{zhu2021data}
Zhuangdi Zhu, Junyuan Hong, and Jiayu Zhou.
\newblock Data-free knowledge distillation for heterogeneous federated learning.
\newblock In \emph{International conference on machine learning}, pages 12878--12889. PMLR, 2021.

\end{thebibliography}
}


\end{document}